\documentclass[runningheads]{llncs}
\usepackage{lmodern}
\usepackage{amsmath}
\usepackage{booktabs}
\usepackage{xcolor}
\usepackage{graphicx}
\usepackage{amssymb}
\usepackage{algpseudocode}
\usepackage{xr-hyper}
\usepackage{hyperref}
\usepackage{algorithm2e}
\usepackage{subcaption,lipsum}
\usepackage{float}
\usepackage{tabularx}
\usepackage{multicol}
\usepackage{multirow}
\usepackage{colortbl}
\newcolumntype{C}{>{\centering\arraybackslash}X}
\usepackage[capitalise,noabbrev]{cleveref}

\makeatletter

\newcommand*{\addFileDependency}[1]{
\typeout{(#1)}
\@addtofilelist{#1}
\IfFileExists{#1}{}{\typeout{No file #1.}}
}\makeatother

\newcommand*{\myexternaldocument}[1]{%
\externaldocument{#1}%
\addFileDependency{#1.tex}%
\addFileDependency{#1.aux}%
}

\myexternaldocument{supp}

\DeclareMathOperator*{\argmax}{argmax}  
\begin{document}
\newcommand{\vbl}[1]{\textcolor{red}{[VBL: #1]}}
\newcommand{\nbr}[1]{\textcolor{green}{[NBR: #1]}}

%
\title{Efficient Precision Control in Object Detection Models for Enhanced and Reliable Ovarian Follicle Counting}
\titlerunning{Efficient Precision control}
%
\author{Vincent Blot\inst{1,2}\and
Alexandra Lorenzo de Brionne\inst{1} \and
Ines Sellami\inst{4} \and Olivier Trassard\inst{6} \and Isabelle Beau\inst{4} \and Charlotte Sonigo \inst{4, 5} \and Nicolas J.-B. Brunel\inst{1, 3}}
%
\authorrunning{V. Blot}

%

\institute{%
Capgemini Invent, 147 Quai du Président Roosevelt, 92130, Issy-Les-MoulineauxFrance\and
Paris-Saclay University, CNRS, Laboratoire Interdisciplinaire des Sciences du Numérique, 91405, Orsay, France \and
Paris-Saclay University, Laboratoire de Mathématiques et Modélisation d'Evry\\ ENSIIE, 1 square de la Résistance, 91000, Évry-Courcouronnes, France \and
Université Paris-Saclay, Inserm, Physiologie et Physiopathologie Endocriniennes,  Le Kremlin-Bicêtre, France \and
AP-HP, Hôpital Antoine Béclère, Service de Médecine de la reproduction et Préservation de la Fertilité, Clamart, France
 \and
INSERM, Institut Biomédical de Bicêtre, Le Kremlin Bicêtre, France}
%
\maketitle              
\begin{abstract}
    Image analysis is a key tool for describing the detailed mechanisms of folliculogenesis, such as evaluating the quantity of mouse Primordial ovarian Follicles (PMF) in the ovarian reserve. The development of high-resolution virtual slide scanners offers the possibility of quantifying, robustifying and accelerating the histopathological procedure.  A major challenge for machine learning is to control the precision of predictions while enabling a high recall, in order to provide reproducibility. We use a multiple testing procedure that gives an overperforming way to solve the standard Precision-Recall trade-off that gives probabilistic guarantees on the precision. In addition, we significantly improve the overall performance of the models (increase of F1-score) by selecting the decision threshold using contextual biological information or using an auxiliary model. As it is model-agnostic, this contextual selection procedure paves the way to the development of a strategy that can improve the performance of any model without the need of retraining it. 

\end{abstract}
\section{Introduction and related works}
\begin{figure}
    \centering
    \includegraphics[width=.75\textwidth]{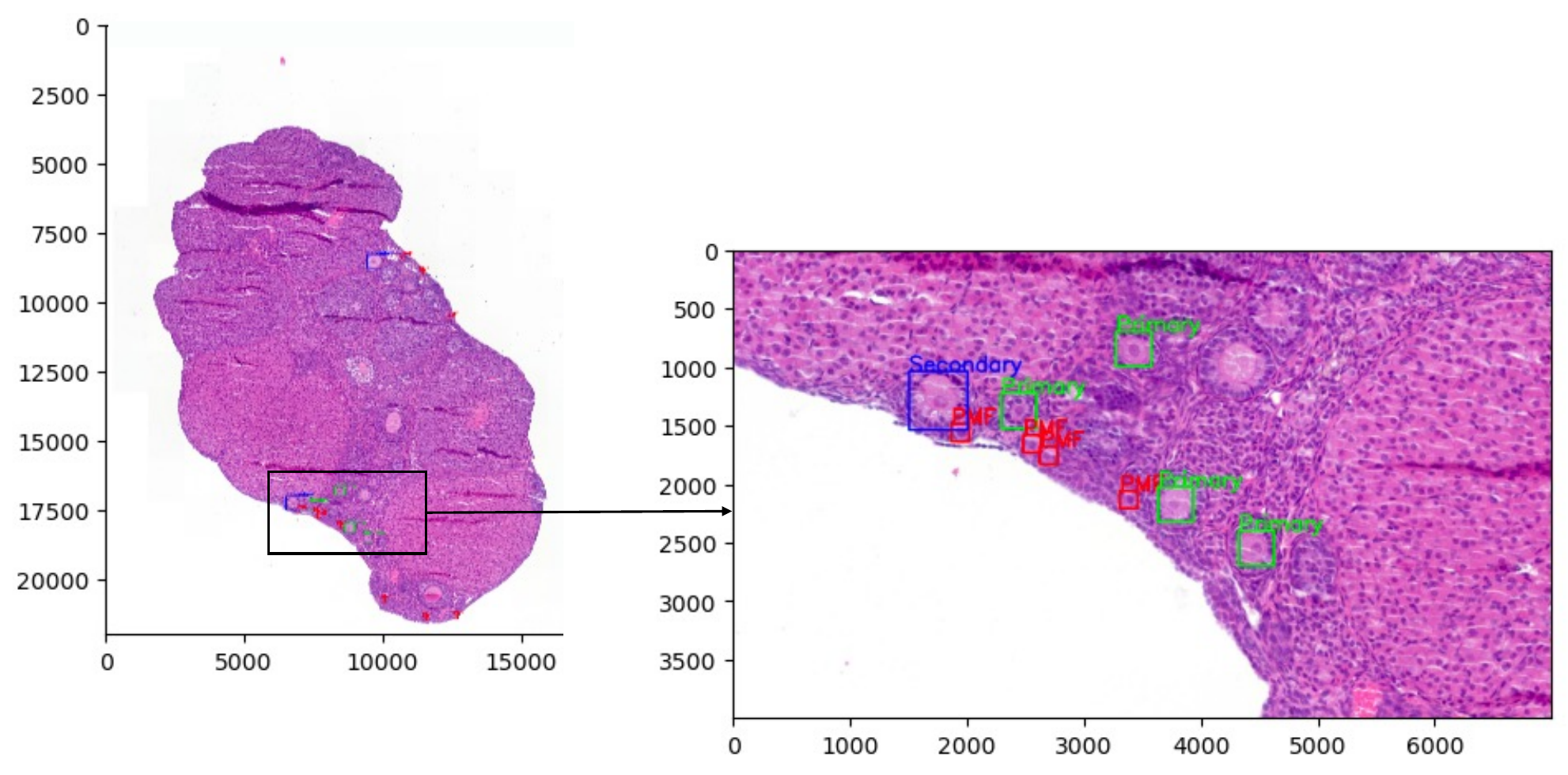}
    \caption{Example of an ovary cut with a zoom on an annotated area with PMF (red), Primary (green) and Secondary (blue) follicles}
    \label{fig:OvaryCut}
\end{figure}

Computer-aided medical image analysis helps improving medical research and clinical practices, in particular for analyzing and describing complex and detailed mechanisms, such as the folliculogenesis. The ovarian follicular stockpile is constituted by primordial follicles (PMF) consisting of an oocyte stuck in the first division of meiosis surrounded by a few somatic cells, known as granulosa cells. These PMFs remain in a quiescent state until initial recruitment, or follicular activation. 
This first step of folliculogenesis is defined by an increase in the size of the oocyte accompanied by a differentiation of the flattened granulosa cells into cuboidal cells. The activated PMFs  have become primary follicles and can undergo the different steps of follicular growth. 
It allows the progression of the primary follicle to secondary follicles then small antral stage~\cite{reddy2016}. Morphologically, it is characterized by a significant increase in the size of the oocyte, differentiation, and proliferation of granulosa cells, and the formation of a liquid-filled cavity at the core of the follicle. Such mechanism can be observed with high-resolution virtual slide scanners, see~\cref{fig:OvaryCut}.


Evaluating the overall count of PMFs within the ovaries provides valuable insights into ovarian reserve and reveals the ultimate impact of certain treatments or pathologies on fertility. Additionally, assessing the total number of primary and secondary follicles helps elucidating the underlying mechanisms affecting the primordial follicular pool and subsequent fertility, such as  the effects of chemotherapies, with high gonadotoxicity on ovaries \cite{Sonigo2019,huang2021}. Eventually, it enables the development of novel therapeutic strategies to limit these ovarian damages. However, identifying and quantifying follicles in mouse ovaries is a time-consuming task, usually carried out manually, which has led to the development of various alternative counting methods~\cite{tilly2003,sarma2020}.\\
Deep Learning is a powerful tool for histological section analysis and histopathology \cite{mehrvar2021deep,cancer}, but it remains underused in fertility studies whereas it could automate the PMFs detection process, thereby facilitating and improving research into reproductive function. A noticeable attempt for counting PMFs in mouse ovaries is proposed in ~\cite{sonigo2018high}, where an ad-hoc convolutional neural network is developed; followed by the use of standard models in  \cite{inik2019new,cai2020automatic,sarma2020}. However, the detection and classification task remains challenging because of the size of the images to be analyzed and subtleties characterizing the follicles.

The aim of the object detection (OD) algorithm is to accelerate the counting of the follicles and to improve its reliability and reproductibility by reducing the intra and inter operator variance. The model should have a high $\mathtt{Recall}$, but also a controlled $\mathtt{Precision}$ in order to give trust in the predictions, as well as reducing the cost of checking the correctness of predictions. A major difficulty is that the uncertainty associated to the selection of the threshold for controlling the precision (or the recall) is rarely taken into account: consequently large variations of performance are often observed. 
In order to solve the standard trade-off, while controlling the uncertainty of the selection process, we maximize the $\mathtt{Recall}$ under the constraint that, with high probability, $\mathtt{Precision}$ is equal to $P_0$, where $P_0$ is user-defined. To achieve this, we leverage recent results from Distribution Free Uncertainty Quantification theory tightly related to the domain of conformal prediction~\cite{gentle_intro_angelopoulos_2023}. Contrary to a straightforward selection of the decision threshold on the objectness, we use a multiple testing procedure introduced in \cite{LTT} in order to select the threshold $\Tilde \lambda$ for detecting objects from the boxes outputted by the model. 
Similar approaches have been successfully introduced in computer vision in order to give probabilistic guarantees on Recall, IOU,\dots \cite{rcps,crc,andeol2023confident}, and are at the basis of Trustworthy Artificial Intelligence. Our main contributions are:

\begin{itemize}
    \item Efficient models for follicle counting accompanied by a guarantee on precision which permits reproducibility of the measurements and reduced human time verification\footnote{Code available at \url{https://github.com/vincentblot28/follicle-assessment}}.
    \item A new open-sourced dataset of high-quality images\footnote{Data available at \url{https://zenodo.org/records/12804564}}.
    \item A contextual-aware object detection procedure that can use biological information and analysis of the errors that gives probabilistic guarantee on the precision, and improve the recall.
\end{itemize}
We emphasize that our context-aware detection procedure is of general interest as it is model-agnostic and can improve the overall performance (F1-score) of any model without retraining it, as the detection rule is directly learned from the data.

\section{Method}

\subsection{Selecting hyperparameters in Object Detection Models} 

We consider that the model $\hat{f}$ is trained to detect objects belonging to $K$ classes. For any image $x \in\mathcal{X}=\mathbb{R}^{N\times N}$, $\hat f(x)= \mathbf{b} \in \mathcal{B}$ is a list of boxes $\mathbf{b} = \{b_1, \ldots, b_{n(x)}\}$ with $b_i = (x_i, y_i, w_i, h_i, s_{i, 1}, \ldots, s_{i, K}, c_i)$ where $(x_i, y_i, w_i, h_i)$ are the coordinates of the box and $s_{i, k}$ is the classification score of the box for the class $k$ and $c_i$ is the objectness (confidence score) of the box. We have a calibration dataset $\mathcal{D} = \{(x_1, \mathbf{b}_1), \ldots, (x_n, \mathbf{b}_n)\}$ that is used for selecting the decision threshold. 
Object detection algorithms are commonly evaluated using metrics such as mean Average Precision ($\mathtt{mAP}$), $\mathtt{Precision}$, and $\mathtt{Recall}$. We denote $\mathtt{Precision}(\mathbf{b},\widehat{\mathbf{b}})$ when we compare the ground truth $\mathbf{b}$ and the prediction $\widehat{\mathbf{b}}$. These metrics provide insights into the performance of the detector. The $\mathtt{mAP}$ is calculated as the mean of the Average Precision (AP) across all classes. AP, in turn, is computed as the area under the precision-recall curve. While $\mathtt{mAP}$ is a global metric over all possible thresholds of objectness, $\mathtt{Precision}$, and $\mathtt{Recall}$ are computed for a specific threshold. The trade-off between $\mathtt{Precision}$, and $\mathtt{Recall}$ can be visualized thanks to the Precision-Recall curve (function $pr: R\mapsto P$).
Follicles are detected in the image $x^{new}$, for each $b_i \in \hat f(x^{new})=\widehat{\mathbf{b}}^{new}$ such that $c_i \geq \lambda$. 
We denote this last layer of the object detection algorithm as the post-processing decision operator $\mathcal{T}_{\lambda}(\mathbf{b})$. The hyperparameter $\Tilde \lambda_0$ is generally selected in order to get $\mathtt{Precision}=P_0$ based on the calibration dataset $\mathcal{D}$, and we derive the corresponding recall $R_0$ as the solution to $R_0 = pr(P_0)$. We call this selection method as naive, as there is no way to assess the generalization power of $\Tilde \lambda_0$: if $(X^{new},\mathbf{b}^{new}) \sim P^{new}$, we have no guarantee that $\mathtt{Precision}(\mathbf{b}^{new},\mathcal{T}_{\Tilde \lambda_0}(\hat{\mathbf{b}}^{new})) = P_0$, nor \[\mathbb{E}_{P^{new}}\left[\mathtt{Precision}(\mathbf{b}^{new},\mathcal{T}_{\Tilde \lambda_0}(\hat{\mathbf{b}}^{new}))\right] = P_0.\]

In the context of controlling the cost associated to the correction of False Positives (proportional to $(1-P_0)+\gamma (1-R_0)$, for $\gamma>0$) or for insuring the reproducibility of the counting process (impacted by the quality of the prediction), we propose a principled way for selecting $\lambda$ such that a minimal expected precision $P_0$ is guaranteed for every new images without degrading the recall too much. As we can see in Fig.~\ref{fig:results} with the red violin plot, the precision with $\Tilde \lambda_0$ can vary a lot above or below $P_0$: while the target precision is $P_0=0.4$, the median is below 0.4, the object detector exhibits significant performance variability, that impacts negatively the reproducibility and trust. As a consequence, we propose to enhance the standard precision control  by introducing a contextual-aware decision operator that can improve the overall performance of the model. This improvement of the performance is possible by leveraging the "Learn Then Test" (LTT) framework~\cite{LTT}, that  provides a probabilistic guarantee on the target precision.


\subsection{Guarantee on the precision of an OD model}

A solution for selecting a proper threshold is to inflate the threshold based on statistical arguments, as it is done in statistical testing for controlling the significativity. For this reason, we use the LTT methodology proposed in \cite{LTT} that sees the selection of the parameter $\lambda$ as a multiple testing problem. \\
We denote for brevity $\mathtt{P}^{(1)}(\lambda) = \mathbb{E}_{P^{new}}[\mathtt{Precision}(\mathbf{b}^{new}$ $,\mathcal{T}^{(1)}_{\lambda}(\hat{\mathbf{b}}^{new}))]$. If we introduce the statistical hypothesis $\mathcal{H}_0^{1,\lambda}: \mathtt{P}^{(1)}(\lambda) \leq P_0$, for any $\lambda$, this means that rejecting the null hypothesis is equivalent to claim that $\lambda$ is \emph{compatible} with the guarantee that the precision of the post-processing $\mathcal{T}^{(1)}_\lambda$ is higher than $P_0$. Instead of considering a continuous set of thresholds $\lambda$, we consider only a discrete set $\lambda_1,\dots, \lambda_m$, and we use multiple testing technics for controlling the False Discovery Rate, i.e. the number of $\lambda_j$s that are wrongly considered to provide a performance $\mathtt{P}^{(1)}(\lambda_j)\geq P_0$.
In order to enable such an approach, it is necessary to have reliable and sharp p-values $p_j$ for each hypothesis $\mathcal{H}_0^{1,\lambda_j}$. As recommended in \cite{LTT}, the test statistic is $\frac{1}{n}\sum_{i=1}^{n}\mathtt{Precision}(\mathbf{b}_i, \mathcal{T}^{(1)}_{\lambda}(\widehat{\mathbf{b}}_i))$ and we use the Hoeffding-Bentkus inequality \cite{rcps} to compute the p-value of the tests $p_j$. Among the approaches introduced for controlling the family-wise error rate (FWER) \cite{dudoit2008multiple}, we choose the so-called Fixed Sequence Testing (FST, see \cite{LTT} for other technics). The methodology is then: 
\begin{enumerate}
    \item Split the dataset into a training set and a calibration set $\mathcal{D}$.
    \item Train the OD model on the training set.
    \item Define a set of thresholds $\Lambda_O = \{\lambda_1, \ldots, \lambda_m\}$, and for each threshold $\lambda_j \in \Lambda_O$, associate the null hypothesis $\mathcal{H}_0^{1,\lambda_j}: \mathtt{P}^{(1)}(\lambda_j) \leq P_0$. In practice, we use $\Lambda_O = \{0.1, 0.2, \ldots, 0.9\}$.
    \item For each null hypothesis $\mathcal{H}_0^{1,\lambda_j}$, compute the p-value $p_j$ based on the Hoeffding-Bentkus inequality (\Cref{eq:hb}).
    \begin{equation}
       p_j^{HB} = \min \left\{\exp \left(-nh_1 \left( \hat{R}_j \wedge \alpha, \alpha \right)\right), eP(Bin(n, \alpha) \leq \lceil n\hat{R}_j \rceil) \right\}
       \label{eq:hb}
    \end{equation}
    \emph{where} $h_1(a, b) = a \log(a/b) + (1 - a) \log((1 - a)/(1 - b))$.
    \item Return $\widehat{\Lambda}_O = A(\{p_j\}_{j \in \{1, \ldots N\}}) \subset \Lambda$, where $A$ is an algorithm that controls the FWER. Here, we use the Fixed Sequence Testing (FST) that gives a set of \emph{compatible} thresholds $\widehat{\Lambda}_O \subset \{\lambda_j : p_j \leq \delta/m'\}$, where $m'<m$ is the number of starting points of the FST algorithm (this algorithm is less conservative than the Bonferroni procedure). 
\end{enumerate}
Once we have the set $\widehat{\Lambda}_O$, we  maximize the mean   $\mathtt{Recall}$ which means that we select $\widehat{\lambda}^*_O=\min \{\lambda_j \in \widehat{\Lambda}_O\}$, because of the 1-to-1 relationship between precision and recall. From Theorem 1 in \cite{LTT}, our selection procedure gives $\hat{\lambda}^*_O = \hat{\lambda}^*_O(\mathcal{D})$, such
that for a user-defined probability $1-\delta>0$, 
\begin{equation}
    \mathbb{P}_\mathcal{D} \Big( \mathbb{E}\left[\mathtt{Precision}(\mathbf{b}^{new},\mathcal{T}_{\hat{\lambda}^*_O}(\widehat{\mathbf{b}}^{new}))\right] \geq P_0 \Big) \geq 1-\delta. \label{eq:Guarantee1}
\end{equation}
where $\mathbb{P}_\mathcal{D}$ is the probability under the calibration dataset.
As we show in section 3, the guarantee~(\Cref{eq:Guarantee1}) is satisfied on the follicule dataset thanks to the more conservative decision threshold selected by the LLT procedure, but at the price of a dramatic drop in the $\mathtt{Recall}$. The aim of the next section is to show that we can reduce significantly this drop, and improve the overall performance as measured by $F1-$score.

\subsection{Improving $\mathtt{Recall}$ with multicriteria decision}
We solve the $\mathtt{Precision}$-$\mathtt{Recall}$ trade-off by replacing the univariate post processing decision operator $\mathcal{T}^{(1)}_\lambda$, with $\lambda \in [0,1]$, by a multiparameter operator $\mathcal{T}^{(2)}_{\lambda,\mu}$, with $\lambda,\mu\in (0,1)$, evaluated on finite grids $\Lambda_O$ and $\Lambda'$. 
The previous approach encompasses such a situation by considering more assumptions $\mathcal{H}_0^{2,\lambda_i,\mu_j}: \mathtt{P}^{(2)}(\lambda,\mu)\leq \mathtt{P}_0$, based on the finite set of couples $(\lambda_i,\mu_j) \in \Lambda_{(2)} = \Lambda_{O} \times \Lambda'$. $\mathtt{P}^{(2)}(\lambda,\mu)$ corresponds to the precision obtained with $\mathcal{T}^{(2)}_{\lambda,\mu}$. We can apply the steps 4-5 of the previous methodology in order to obtain the set of \emph{compatible} thresholds $\widehat{\Lambda}_{(2)}$, that corresponds to the set of hyperparamaters $(\lambda,\mu)$ such that $\mathtt{P}^{(2)}(\lambda,\mu)\geq \mathtt{P}_0$, and we select the decision thresholds as
\begin{equation}
    (\widehat{\lambda}^*_{(2)}, \widehat{\mu}_{(2)}^*) = \argmax_{(\lambda, \mu) \in \hat{\Lambda}_{(2)}} \frac{1}{n}\sum_{i=1}^{n} \mathtt{Recall} (\mathbf{b}_i,\mathcal{T}^{(2)}_{\hat{\lambda}, \hat{\mu}}(\widehat{\mathbf{b}}_i)).
\end{equation}
As $\Lambda_{(2)}$ is bigger than $ \Lambda_O$ (that is included in $\Lambda_{(2)}$ with a default parameter $\mu$), we can reach higher values for the $\mathtt{Recall}$ by optimising on $\widehat{\Lambda}_{(2)}$, while maintaining the guarantee on $\mathtt{Precision}$. This can be extended straightforwardly to $d$ hyperparameters. \\
Hence the efficiency of our approach relies on the ability to design sharp post processing operators. We describe below two generic ways of building sharper decision operators based on the analysis and the reduction of the errors generated by the simple decision rule $c_i > \lambda$. We introduce below two families of extra criterion for predicting a bounding box.


\subsubsection{Exploitation of interpretable biological information}
\begin{figure}
    \centering
    \includegraphics[width=.4\textwidth]{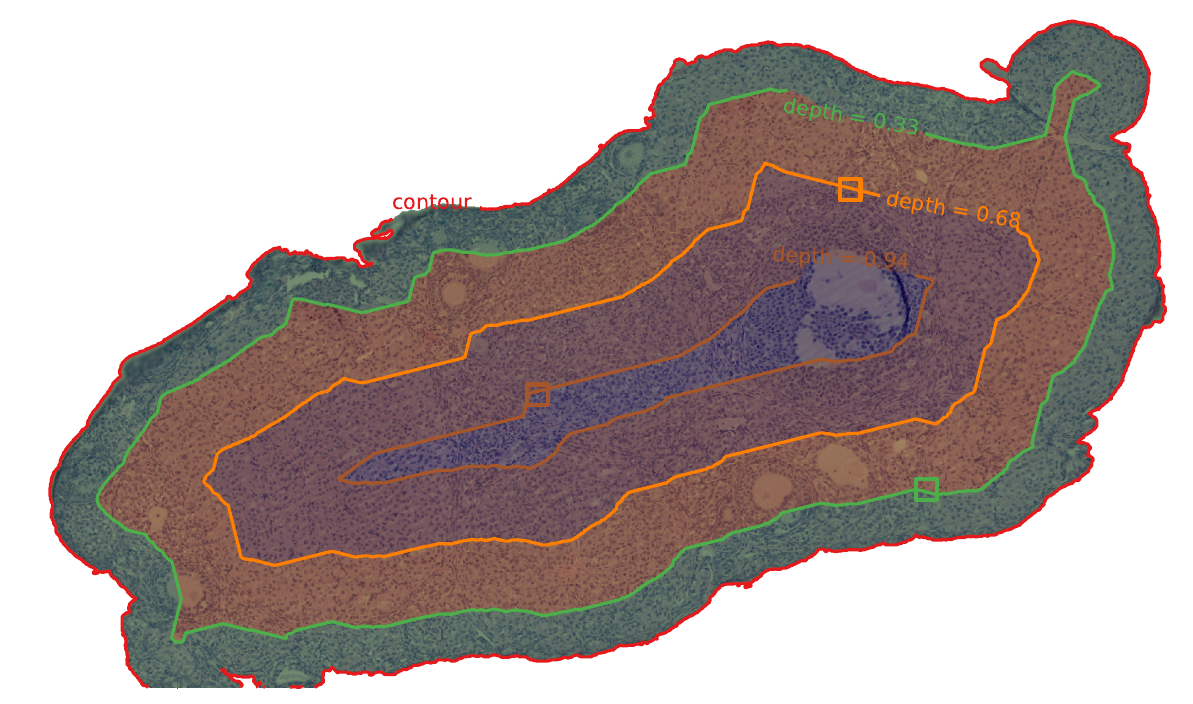}
    \caption{Computation of the depth of a box. The box is predicted by the OD model and the contour of the ovary is computed. The contour is then dilated until the box is inside. The depth of the box is then computed as the ratio of the area of the dilated contour over the area of the ovary. Each line represents the contour-line of the dilatation associate to a bounding-box and each box represents a prediction of the model.}
    \label{fig:depth}
\end{figure}

We start by considering "use-case specific" information coming from biological expertise. Such information are often available in practice, but there are difficult to exploit in Deep Learning/Computer Vision algorithms and they are often used as probes for post-evaluation of the quality of prediction.\\
In our case, we know that folliculogenesis process is mainly located at the periphery of the ovary, meaning that we cannot expect a lot of follicles deep inside the ovary (see \Cref{fig:depth_dist} in supplementary material). For this reason, we introduce the \emph{depth} $d_i=D(b_i)$ of a box $b_i$ (or the distance with respect to periphery), such that  $d_i=1$ if $b_i$ is predicted exactly at the center of the ovary, and $d_i=0$ if $b_i$ is predicted on the boundary (see Figure~\ref{fig:depth}). Our definition of depth is based on the detection of the contour but in order to reduce the sensitivity to the quality of the detection of the contour, and solve the problem of defining a unique distance between an inner point and the contour (because the follicules are non-convex in general), we use the inner volume between the contour an the inner point. The exact computation of the depth $D(\cdot)$ is given in the supplementary material~(\Cref{alg:depth}). 

We introduce then the two-parameter post-processing detector $\mathcal{T}_{(\lambda, \mu)}(\mathbf{b}) = \{ b_i : c_i \geq \lambda \;\textrm{and}\; d_i \leq \mu\}$, that will predict a box if the objectness is high enough, and also if it is not too deep. The previous general methodology can be applied for selecting the appropriate thresholds $(\widehat{\lambda}^*_{O,D}, \widehat{\mu}_{O,D}^*)$.

\subsubsection{Prediction of False Detections.} The previous approach proposes an interpretable analysis and correction of the errors of the object detector, but in general we can explain the errors of the model by learning the false detections. We can learn to classify the falsely detected boxes and actual detected boxes, i.e., we predict 1 if the rule $c_i > \lambda$ provides a good detection and 0 if not. It is often the case that it remains some signal in the data that can be exploited. There is some parallel with boosting~\cite{boosting_schapire}, where a sequence of models is used to improve the residuals of the previous ones. In our case, instead of optimising only the performance (such as accuracy), we maximize the $\mathtt{Recall}$ under a probabilistic guarantee on the $\mathtt{Precision}$. To do this, once the OD model trained, we run an inference on all the training data, and for each predicted bounding box, we create fixed size patches around it, and assign them the label 1 if there is an follicle inside the box and 0 otherwise (we do not consider the class predicted in the OD, but only presence or not of the box). 
We train a CNN $g$ for this binary classification task on the same training data as the OD. The final post processing detector is defined as $\mathcal{T}_{(\lambda, \mu)}(\mathbf{b}) = \{ b_i : c_i \geq \lambda \;\textrm{and}\; g(b_i) \geq \mu\}$ ($g(\cdot)$ is the probability of true detection). The hyperparameter selection is done in $\Lambda_{O} \times \Lambda_{Classif}$, where $\Lambda_{Classif}$ is a discretized grid in $(0,1)$, and we obtain $(\widehat{\lambda}^*_{O,Classif}, \widehat{\mu}_{O,Classif}^*)$.

\section{Experimental setup \& Results}


\subsection{Dataset}

Our dataset is composed of 643 cut slices of ovaries coming from 92 mice (each mouse have between 6 and 8 cuts). Ovaries were fixed in Bouin's solution and subsequently embedded in paraffin blocks. The entire ovaries were serially sectioned into $4\mu m$ slices using a microtome. Every fifth section was mounted on microscope slides and stained with Hematoxylin and Eosin. Ovaries have been digitized on a Panoramic 250 Flash, Slide Scanner (3DHISTECH Ltd. HUNGARY). Each section was analyzed using Calopix Viewer®. Follicles were classified according to Pedersen’s \cite{pedersen1968proposal} classification. Briefly, follicles were categorized as follows: primordial follicles contained an oocyte surrounded by a partial or complete layer of squamous granulosa cells; primary follicles had an oocyte surrounded by a single layer of cuboidal granulosa cells; secondary follicles presented with at least two layers of granulosa cells; and antral follicles featured a visible antrum cavity.
The size of those cuts is approximately $20000 \times 20000$. As cuts coming from the same mouse can be quite similar, we have decided to perform the split between train, calibration and test set at the mouse level to avoid potential overfitting (i.e. two cuts coming from the same mouse can't be into two different sets). For training our algorithm, we used 503 cuts and the rest will be used for calibration and test.
As the cuts images are way to large to be processed by a neural network, we split them into patches of size $1000 \times 1000$ with a stride of $500$. We choose to keep only $5\%$ of patches without follicle, resulting in a dataset with 31256 images, among which 16245 contain at least one labelled follicle. This dataset was separated into training and validation with a 85\%/15\% split). For the calibration, as we want to achieve a precision control at the cut level (and not the patch), we consider only the cut images (and the patches will be used for the prediction of our OD model). We have 35 ovary cuts for the calibration and 35 for test with 12 annotations by cut on average.

\subsection{Training and results of the OD and the classification models}

We ran our experiments with two different models : EfficientDet~\cite{EfficientDet} and YoloV8~\cite{Jocher_Ultralytics_YOLO_2023}. Performance of both detectors is displayed in~\cref{tab:ths_ltt} (a detailed version is availble in~\cref{tab:pr_all} of supplementary material). For both OD models, the auxiliary classification model is a VGG16~\cite{vgg} network pre-trained on the ImagetNet dataset~\cite{imagenet}. It achieves 75\% accuracy on the validation dataset. Training parameters for all three models are described in~\cref{tab:training_params_comparison} of the supplementary materials.
\subsection{Results of our methodology}
\begin{figure}[h!]
    \tiny
    \centering
    \includegraphics[width=1\linewidth]{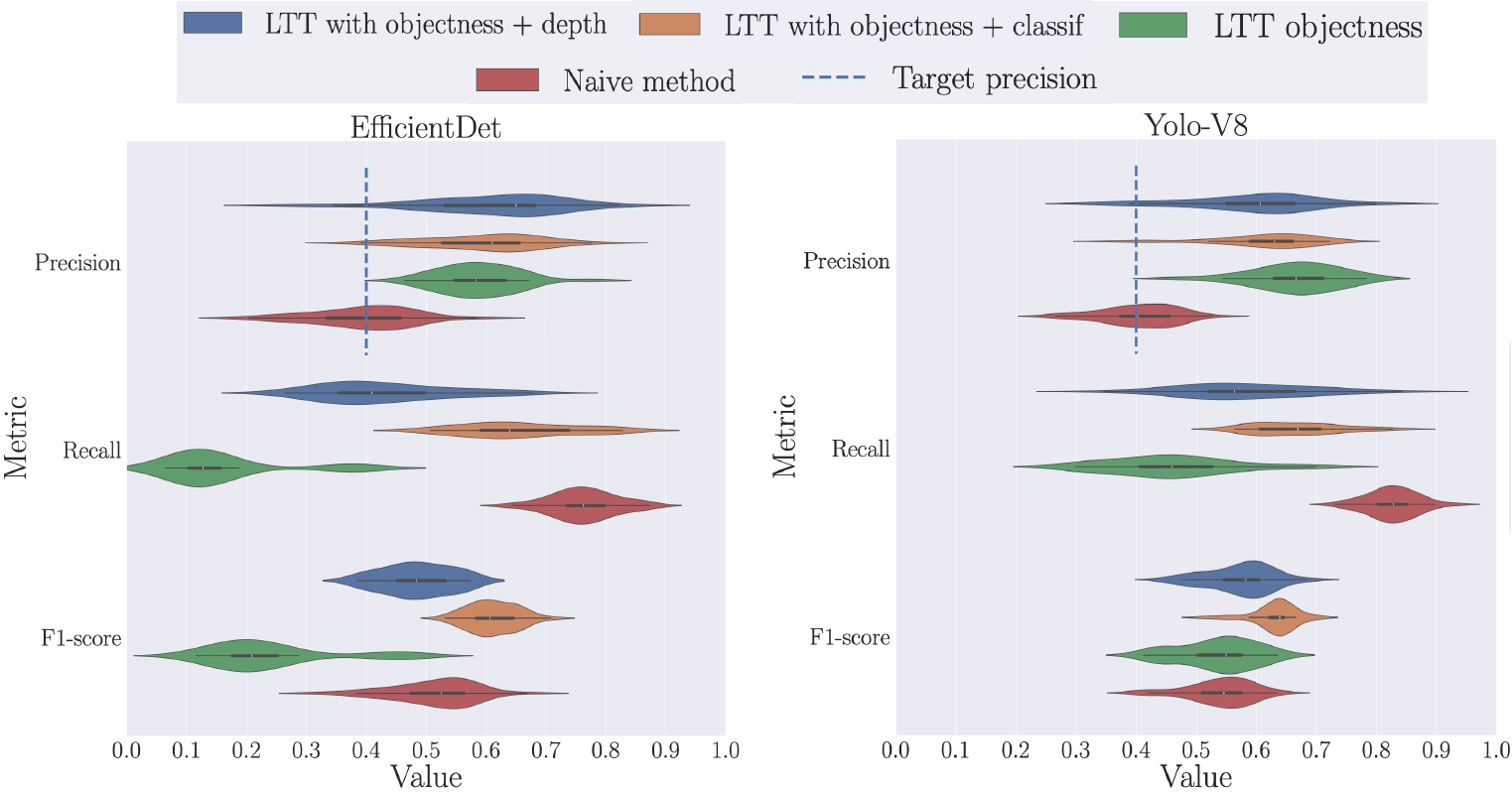}
  
  \caption{Precision, Recall and F1-score for target precision $P_0 = 0.4$. \textbf{Left:}  EfficientDet model. \textbf{Right}: Yolo model. In blue: decision with objectness and depth; in orange: decision with objectness and classification score, in green: LTT decision with  objectness only; in red: the naive decision.}
  \label{fig:results}
\end{figure}

The results of our methodology are shown in~\cref{fig:results}. Those results were generated over 100 independent data splits between calibration and test. The  thresholds $\Tilde \lambda_0, \widehat{\lambda}^*_O, \left(\widehat{\lambda}^*_{O,D},\widehat{\mu}^*_{O,D}\right)$ and $\left(\widehat{\lambda}^*_{O,C},\widehat{\mu}^*_{O,C}\right)$ are computed for a target precision $P_0=0.4$ and $\delta = 10^{-3}$. The thresholds returned by the LTT procedure are displayed in~\cref{tab:ths_ltt}.

First, we can see that either with  LTT on objectness or  LTT on depth and objectness, we achieve a  control of the Precision above with a high probability, contrary to the naive method where almost half of the observation have a precision lower than expected. Moreover, our methodology achieves a higher F1-score than LTT on a single parameter; and a higher recall than both the "standard LTT" and the naive method, which makes it more efficient than those methods (because with the same guarantee on the precision, we do improve the F1-score).

\begin{table}[ht!]
    \centering
    \tiny
    \caption{Performance of the OD models and thresholds returned by the LTT procedure and the naive method for EfficentDet and Yolo models}
    \begin{tabularx}{\linewidth}{ccCc|cCc}
        \toprule
        & \multicolumn{3}{c|}{\textbf{EfficientDet}} & \multicolumn{3}{c}{\textbf{YOLO-V8}} \\
        \midrule
        & \textbf{Precision (\%)} & \textbf{Recall~(\%)} & \textbf{ mAP (\%)} & \textbf{Precision (\%)} & \textbf{Recall~(\%)} & \textbf{ mAP~(\%)} \\
        \multirow{-2}{*}{All Classes} & 29.8 & 83.8 & 32.8 & 44.2 & 74.9 & 33.7   \\
        \cmidrule{1-7}
        \textbf{Thresholds} & \textbf{Objectness} & \textbf{Depth} & \textbf{Classification} & \textbf{Objectness} & \textbf{Depth} & \textbf{Classification} \\
        \cmidrule{1-7}
        \rowcolor{gray!25}
        Naive Method & 0.568 & - & - & 0.405 & - & - \\
        LTT Objectness & 0.700 & - & - & 0.699 & - & - \\
        \rowcolor{gray!25}
        LTT Objectness + Depth & 0.642 & 0.515 & - & 0.618 & 0.514 & -  \\
        LTT Objectness + Classification & 0.534 & - & 0.230 & 0.461 & - & 0.214  \\
        \bottomrule
    \end{tabularx}\label{tab:ths_ltt}

\end{table}

\section{Discussion and Conclusion}
We have introduced a general methodology that permits to control the uncertainty of the hyperparameter selection process in a more efficient way. As a consequence, we have accelerated and robustified follicle counting. Such an increase in reproducibility will contribute to the adoption of AI among physicians, and ability to compare clinical studies. As our approach is model-agnostic and relies on a scalable multiple testing procedure, it shows that repeatability and performance can be both augmented by abandoning the naive threshold selection. Future work will focus on refining multiple testing strategies for improving this gain, and in defining and learning more complex post processing detectors, with various models and biological information.

\newpage
\bibliographystyle{splncs04}
\bibliography{miccai-ltt.bib}

\end{document}